# Why is language well-designed for communication?

Jean-Louis Dessalles
*Telecom ParisTech*
**www.dessalles.fr**

**Abstract:** Selection through iterated learning explains no more than other non-functional accounts, such as universal grammar, why language is so well-designed for communicative efficiency. It does not predict several distinctive features of language like central embedding, large lexicons or the lack of iconicity, that seem to serve communication purposes at the expense of learnability.

Christiansen and Chater rightfully observe that communicatively arbitrary principles, such as UG, are unable to explain why language is adequate for communication. The same criticism can be addressed, however, to their own account. If the main driving force that led to language emergence is learnability rather than communicative efficiency, language should be locally optimal for the former and *not* for the latter. Evidence suggests that, in several respects, the exact opposite is the case.

What would language be like if, as the authors claim, the cultural selection of learnable languages were "stronger" than the biological selection of brains designed for efficient communication? If language can compare with a "viral" entity that gets selected for its ability to resist vertical cultural transmission, we predict for instance *iconic signifiers*, especially gestures, to win the contest. Yet, although analogue resemblance makes learning almost trivial, linguistic evolution shows that non-iconic signifiers tend to prevail, even in sign languages.

The "viral" theory of language does not explain the *size of lexicons* either. Ideally, an expressive code is easiest to learn, and resists iterated transmission best, if words are limited in number and have separate and unambiguous meanings. Yet, real vocabularies include tens of thousands of words, massive near synonymy and many rare unpredictable word combinations (Briscoe, 2006). Such evidence suggests that there may be some "viral" cause for the existence of plethoric lexicons, but its action is opposite to what is expected from selection for learning efficiency.

Language, as mainly shaped by selection through repeated learning, is supposed to mirror the general human induction bias. Efficient induction systems (Solomonoff, 1978), including human learning (Chater, 1999) and analogy making (Cornuéjols, 1996) are guided by a *complexity minimization principle*. If languages were the bare expression of a simplicity-based induction device looping on itself, we should expect the complexity of languages to converge to a minimal amount. A similar claim is that general-purpose learning devices, except in rote learning mode, produce only "good shapes" (Gestalten), *i.e.* structures that are left invariant by operations forming an algebraic group (Dessalles, 1998a). Language has not, so far, been described as involving good shapes. For instance, syntactic structures, contrary to many other aspects of cognition, cannot be induced as invariants of transformation groups (Piattelli-Palmarini, 1979) and seem to thwart general inductive processes (Piattelli-Palmarini, 1989).

In a bio-functional account of language emergence, learnability puts limits on what is admissible, but is subordinate to communicative functions. The two main proximal functions of language in our species, as revealed by the observation of spontaneous language behavior, are *conversational narratives* and *argumentative discussion* (Bruner, 1986; Dessalles, 2007). From a bio-functional perspective, iconicity is dispensable if the problem is to express predicates for argumentative purposes (Dessalles, 2007). Lexical proliferation is predicted if the problem is to signal unexpectedness in narratives and to express nuances in argumentative discussion (Dessalles, 2007). And language-specific learning bias is expected if early language performance makes a biological difference. Let us consider a fourth example to show that functional aspects of language could evolve at the expense of learnability.

Non-functional accounts of language, including cultural selection through iterated learning, do not account for the existence of *central embedding* (the fact that any branch may grow in a syntactic tree), a feature present in virtually all languages. Recursive syntax has been shown to emerge through iterated learning, but only when individuals already have the built-in ability to use recursive grammars to parse linguistic input (*e.g.* Kirby, 2002). A bio-functional approach to language provides an explanation for the presence of central embedding in language. As soon as the cognitive ability to form predicates is available, possibly for argumentative purposes (Dessalles, 2007), predicates can be recruited to determine the arguments of other predicates. This technique is implemented in computer languages such as Prolog. To express "Mary hit Paul" for listeners who do not know Mary, the speaker may use "Mary ate with us yesterday" to determine the first argument of "hit". Prolog achieves this through explicit variable sharing, whereas human languages connect phrases for the same purpose: "The girl who ate with us yesterday hit Paul" (Dessalles, 2007).

Predicates $P_{1i}$ can therefore be used to determine arguments in a given predicate $P_1$; but each $P_{1i}$ may require further predicates $P_{1ij}$ to determine its own arguments. This possibility leads to recursive syntactic processing that produces central embedded phrase structures. Models that ignore functions like predicate argument determination cannot account for the necessity of embedded phrase processing. They merely postulate it, either as a consequence of some fortuitous genetic accident (Chomsky 1975) or deduce it from a general cognitive ability to perform recursive parsing (Kirby, 2002). But then, the adequacy to the function is left unexplained as well. No single genetic accident and no selection through repeated learning can predict that phrase embedding will efficiently fulfill predicate argument determination. Only a bio-functional approach that derives the existence of phrase embedding from its function can hope to explain why recursive processing came to exist *and* why it is locally optimal for that function.

From a phylogenetic perspective we may wonder why, if human languages have been selected to be easily learned, chimpanzees are so far from acquiring them, spontaneously or not. One must hypothesize some yet unknown qualitative gap between animal and human *general* learning abilities. Invoking such "pre-adaptation" remains, for now, non-parsimonious. Not only is the emergence of "pre-adaptations" not accounted for in iterated learning models and more broadly in non-functional models, but their subsequent assemblage into a functional whole remains mysterious as well. Bio-functional approaches to language emergence avoid the "pre-adaptation" trap. They do not attempt to explain why a given feature did *not* occur in other lineages by invoking the lack of required "pre-adaptations".

Language is not a marginal habit that would be incidentally used in our species. It has dramatic influence, not merely on survival but on differential reproduction, which is what determines natural selection. Individuals who fail to be relevant are excluded from social networks and become preferential victims (Dessalles 1998b; 2007). Given the crucial impact of conversational performance on reproductive success, it would be highly unlikely that human brains could have evolved independently from language.

**References**

Briscoe, T. (2006). Language learning, power laws, and sexual selection. In A. Cangelosi, A. D. M. Smith & K. Smith (Eds.), *The evolution of language*. Singapore: World Scientific.

Bruner, J. (1986). *Actual minds, possible worlds.* Cambridge, MA: Harvard University Press.

Chater, N. (1999). The search for simplicity: A fundamental cognitive principle?. *The Quaterly Journal of Experimental Psychology*, *52* (A), 273-302.

Chomsky, N. (1975). *Reflections on language.* New York: Pantheon Books.

Cornuéjols, A. (1996). Analogie, principe d'économie et complexité algorithmique. In *Actes des 11èmes Journées Françaises de l'Apprentissage*. Sète. Available at: www.lri.fr/~antoine/Papers/JFA96-final-osX.pdf

Dessalles, J-L. (1998a). Limits of isotropic bias in natural and artificial models of learning. In G. Ritschard, A. Berchtold, F. Duc & D. A. Zighed (Eds.), *Apprentissage : Des principes naturels aux méthodes artificielles*, 307-319. Paris: Hermès. Available at: www.dessalles.fr/papiers/pap.cogni/Dessalles_97062502.pdf

Dessalles, J-L. (1998b). Altruism, status, and the origin of relevance. In J. R. Hurford, M. Studdert-Kennedy & C. Knight (Eds.), *Approaches to the evolution of language: social and cognitive bases*, 130-147. Cambridge: Cambridge University Press. Available at: www.dessalles.fr/papiers/pap.evol/Dessalles_96122602.pdf

Dessalles, J-L. (2007). *Why we talk - The evolutionary origins of language.* Oxford: Oxford University Press. www.dessalles.fr/WWT/

Kirby, S. (2002). Learning, bottlenecks and the evolution of recursive syntax. In T. Briscoe (Ed.), *Linguistic evolution through language acquisition: Formal and computational models*, 173-203. Cambridge, MA: Cambridge University Press.

Piattelli-Palmarini, M. (1979). *Théories du langage - Théories de l'apprentissage.* Paris: Seuil.

Piattelli-Palmarini, M. (1989). Evolution, selection and cognition: From 'learning' to parameter setting in biology and in the study of language. *Cognition*, *31* (1), 1-44.

Solomonoff, R. J. (1978). Complexity-based induction systems: Comparisons and convergence theorems. *IEEE transactions on Information Theory*, *24* (4), 422-432. Available at:
world.std.com/~rjs/solo1.pdf